%% file: paper.tex
\documentclass[journal]{vgtc}                

\ifpdf
  \pdfoutput=1\relax                   
  \usepackage{graphicx}                
  \DeclareGraphicsExtensions{.pdf,.png,.jpg,.jpeg} 
\else
  \usepackage{graphicx}                
  \DeclareGraphicsExtensions{.eps}     
\fi%

\graphicspath{{figures/}{pictures/}{images/}{./}} 

\usepackage{microtype}                 
\PassOptionsToPackage{warn}{textcomp}  
\usepackage{textcomp}                  
\usepackage{mathptmx}                  
\usepackage{times}                     
\usepackage{cite}                      
\usepackage{tabu}                      
\usepackage{booktabs}                  
\usepackage{import}
\usepackage{algpseudocode}

\onlineid{1258}

\vgtccategory{Research}
\vgtcpapertype{Algorithm/Technique}

\title{VINE: Visualizing Statistical Interactions in Black Box Models}

\author{Matthew Britton}
\authorfooter{
\item
 Matthew Britton is with the Georgia Institute of Technology. E-mail: matt.britton@gatech.edu.
}

\abstract{As machine learning becomes more pervasive, there is an urgent need for interpretable explanations of predictive models. Prior work has developed effective methods for visualizing global model behavior, as well as generating local (instance-specific) explanations. However, relatively little work has addressed regional explanations - how groups of similar instances behave in a complex model, and the related issue of visualizing statistical feature interactions. The lack of utilities available for these analytical needs hinders the development of models that are mission-critical, transparent, and align with social goals.  We present VINE (\textbf{V}isual \textbf{IN}teraction \textbf{E}ffects), a novel algorithm to extract and visualize statistical interaction effects in black box models. We also present a novel evaluation metric for visualizations in the interpretable ML space.
} 


\CCScatlist{ 
 \CCScat{K.6.1}{Management of Computing and Information Systems}%
{Project and People Management}{Life Cycle};
 \CCScat{K.7.m}{The Computing Profession}{Miscellaneous}{Ethics}
}

\teaser{
  \centering
  \includegraphics[width=\linewidth]{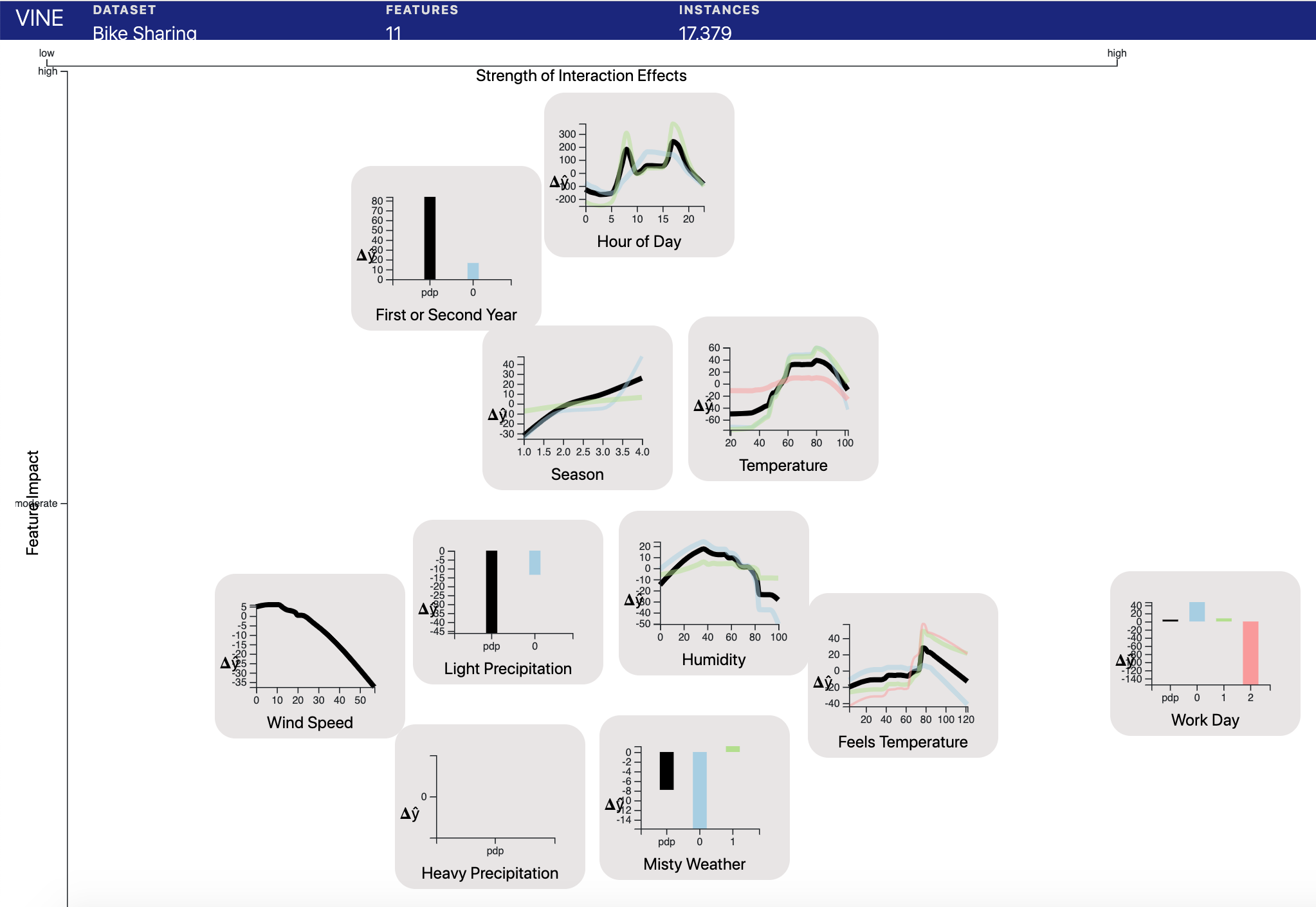}
  \caption{VINE overview screen showing feature space visualization. VINE plots are displayed as small multiples. Position on the X-axis denotes strength of interaction effects, and position on the Y-axis denotes feature importance. Users can click on a chart to open a detailed view with regional model explanations.}
	\label{fig:teaser}
}

\renewcommand{\manuscriptnotetxt}{}

 \vgtcinsertpkg

\begin{document}


\firstsection{Introduction}

\maketitle

\import{sections/}{introduction}

\import{sections/}{background}

\import{sections/}{relatedWork}

\import{sections/}{approach}

\import{sections/}{implementation}

\import{sections/}{evaluation}

\import{sections/}{results}

\import{sections/}{discussion}

\import{sections/}{futureWork}

\import{sections/}{conclusion}

\acknowledgments{The authors wish to thank Fred Hohman and Andrea Hu.}

\bibliographystyle{template/abbrv-doi}
\nocite{*}
\bibliography{references}
\end{document}

%% file: sections/introduction.tex
State-of-the-art machine learning algorithms such as neural networks and support vector machines have shown enormous success in modeling complex data. These models are widely regarded as ``black box'' algorithms, meaning that the reasons why they make a prediction are not clear. There are serious downsides to employing predictive models whose behaviors are not fully understood. A well known example of this hazard was a study of how machine learning could be used to predict pneumonia risk \cite{caruana2015intelligible, cooper2005predicting}. A rule-based, interpretable model extracted a counter-intuitive result from the dataset - having asthma was found to be a protective factor (i.e. it lowered risk). However, asthma is actually an aggravating factor, a fact so well-known by doctors that patients with the condition typically received aggressive treatment, improving their outcomes. If the authors had instead utilized their black box neural network trained on the same data, this behavior would have gone unnoticed and the model would have placed asthmatics lower in the triage order, potentially costing lives.

A case can be made for both social \cite{rudin2018please} and economic \cite{sculley2014machine} gains that would be realized with a shift towards interpretable models. The movement towards this lofty goal was accelerated recently with the passage of the European Union's General Data Protection Regulation \cite{eu-gdpr}, which grants individuals the ``right to an explanation'' for some automated decisions. In this environment, model explanations should be seen as both a functional safeguard and an obvious practicality. 

Taking the need for explainability as a given, we are still faced with major questions regarding implementation. What format should explanations be presented in, such as via summary statistics or generated natural language explanations? Visual analytics systems that present data in a graphical, manipulable format have proven effective for many forms of data analysis, and so it is reasonable to investigate their utility for user exploration of ML models as well. 

What, exactly, should an explanation explain, given that there are a massive number of behaviors that could conceivably be extracted from a complex model, and that data scientists prefer different forms of explanation depending on the problem context \cite{hohman2019gamut}? Existing methods for generating explanations can be generally separated into \textit{global explanations} and \textit{local explanations}. \textit{Global explanations} aim to describe how a model works in broad terms, for example by listing the top \textit{k} most important features. For example, take the task of predicting daily ridership for a bike-sharing program, based on features such as weather, day of the week, etc.. A global explanation might communicate to the user that warm weather increases ridership. A \textit{Local explanation}, on the other hand, typically focuses on a single data point which has been selected by a process external to the model. For example a user may want an explanation for why they were denied a loan, or an engineer may want to investigate an unexpected prediction. Local explanations often take the form of a counterfactual or what-if statements, such as ``ridership would have been 10\% higher if it hadn't rained.''

Both global and local explanations can give insight into a model. However, they also have clear downsides. Global explanations are by necessity simplifications of the real behaviors that the model performs. Higher temperatures are correlated with increased ridership, but perhaps temperature is irrelevant on the subset of days with pouring rain. Local explanations typically have unclear generalizability - a user does not know if the features that were most important in this prediction will be impactful for other subsets of the data. For a local explanation such as ``the temperature for this day (75$^\circ$) increased ridership by 60 over the average)'' it is impossible to gain an understanding of how to interpret temperature's role in the prediction for another case. 

There is still a wide gap between the information conveyed by state-of-the-art explanation methods, and the diverse array of behaviors that a complex predictive model can evince. To address these limitations, we explore a relatively neglected third approach, \textit{regional explanations}. A regional explanation describes how an important subset of instances is treated by a model -  this behavior is of particular interest when the regional explanation differs from the global explanation, so the regional explanation can be presented alongside a global one. For example, the global explanation ``temperature increases ridership'' could be complemented by the regional explanation ``on weekends, temperature increases ridership, but on weekdays, temperature plays less of a role.'' These subsets give insights into the distinctions that a model finds salient, which can then be compared to human intuition.

We present VINE (Visual INteraction Effects), a visualization which highlights unusual or noteworthy model behaviors which only apply to a subset of data cases. We base our approach on partial dependence \cite{friedman2001greedy}, a common method for analyzing the relationship between a feature and model predictions. Relevant subsets are automatically extracted by our algorithm. In additional to a visual depiction of how the model generates a prediction for these subsets, our tool generates a predicate to describe the points in this cluster. The result is a regional explanation that adds nuance to the global explanation.

Finally, how do we evaluate the effectiveness of a particular model visualization? Visualizations in this space are difficult to evaluate, partly because they typically combine a novel algorithm for behavior extraction with a novel visualization. Little work has been done to disentangle the effects of the two. As a result, most evaluations in this space have either employed example use cases constructed by the authors, or purely qualitative evaluations of human participants \cite{doshi2017towards}. While these clearly have their place, the field suffers from a paucity of metrics that allow explanations to be compared quantitatively. To this end, we introduce the Information Ceiling framework, a simple process that attempts to make new predictions based on the information presented in the visualization. By employing the algorithm to compare VINE with partial dependence plots, we find that our visualization outperforms existing methods on multiple datasets. More importantly, we are able to compare visualizations using a simple metric that quantifies their fidelity to the underlying model. 

Our contributions in this paper can be summarized as follows:

\begin{itemize}
    \item A novel algorithm for generating regional explanations of predictive models, based on clustered partial dependence curves;
    \item An interactive visualization, VINE, which allows the user to explore these explanations;
    \item The Information Ceiling framework, a method for evaluating the fidelity of any visual model explanation to the underlying model.
\end{itemize}

%% file: sections/background.tex
\section{Background}

Our approach is heavily based on existing work \cite{friedman2001greedy, goldstein2015peeking} on partial dependence plots and other visualizations that use the partial dependence calculation to generate explanations. The implementation and usage of these plots is described below.

\subsection{Partial Dependence Plots}

\begin{figure}
\centering
\includegraphics[scale=0.55]{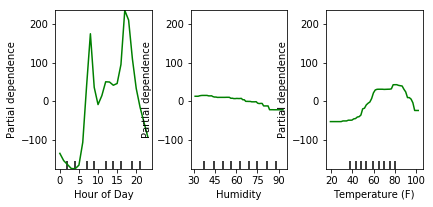}
\caption{Three PDP plots from the bike dataset, showing the main effects for salient features.}
\label{fig:pdp-background}
\end{figure}

Partial Dependence Plots (PDPs) \cite{friedman2001greedy} calculate the average prediction across all instances as the value of a single feature is changed, holding all other values constant. PDPs are typically constructed for each feature in a dataset - a sample of PDPs for the bike dataset are presented in Figure \ref{fig:pdp-background}. Intuitively, the partial dependence curve shows the best guess for a prediction if only one feature value is known. Figure \ref{pdp_equation} is used to generate a partial dependence curve for a single predictor feature:

\begin{equation}\label{pdp_equation}
	pdp_f(v) = \frac{1}{N}\ \sum_i^N pred(x_i) \ with \ x_{i_f} = v
\end{equation}

\textit{N} is the number of items in the dataset, \textit{pred} is the function defined by the predictive model, \textit{f} is the predictor feature in question, and \textit{v} is a value in the domain of \textit{f}. The model is treated as an oracle and generates \textit{N} curves constructed of \textit{M} data points each, where \textit{M} is a hyperparameter that determines the granularity of the explanation. \textit{v} takes on the values of the \textit{M} quantiles of \textit{f}.

\begin{figure}
\centering
\includegraphics[scale=0.55]{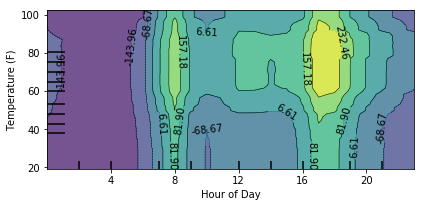}
\caption{2-D PDP plots show interactions and main effects for two features. This example shows the interaction between Temperature and Hour of Day for the bike dataset.}
\label{fig:2d-pdp-background}
\end{figure}

Partial Dependence is one of the most common ways to communicate how a prediction depends on a single feature. The plots are easy to calculate and interpret, and have become fixtures in open-source \cite{austin_2018,scikit-learn,biecek2018dalex} and proprietary \cite{h2o2019} data-science toolkits. In addition to the traditional line chart, they have also been presented as colored bars \cite{krause2016interacting} and 2-feature heatmaps and contour plots (see Figure \ref{fig:2d-pdp-background}). Note that PDPs (and other plots in this family) can be presented with the standard scale (in which the Y-axis is read as the predicted value) or as a centered PDP (in which case the Y-axis is read as the change from the average prediction). Figures \ref{fig:pdp-background} and \ref{fig:2d-pdp-background} present the standard scale, however we use the centered PDP in the rest of this work.

However, PDPs have three main shortcomings:
\begin{itemize}
  \item \textbf{Unreasonable assumption of feature independence.} The synthetic data generated by PDPs may be highly unlikely under the joint distribution, if the input features are correlated. For example, in a dataset of personal health records, predictions would be generated for children up to the maximum height in the dataset, perhaps 6' tall. The predicted target might be outlandish, and skew the summary curve in regions with low probability mass.
  \item \textbf{Heterogeneous effects are obscured by the summary curve.} The process of averaging the curves produced for each data point necessarily obscures varying shapes.
  \item \textbf{Feature interactions are difficult to separate from the main variable effect.} The PDP curve includes all feature interactions, making it difficult to isolate the importance of the feature of interest itself.
\end{itemize}

\subsection{Individual Conditional Expectation plots}

To address PDPs' tendency to obscure heterogeneous effects, \cite{goldstein2015peeking} presented Individual Conditional Expectation (ICE) plots, which disaggregate the PDP line into its constituent curves, one for each data point in the original dataset. The ICE plot consists of a line plot with one series of predictions for each instance. While ICE plots display the full heterogeneity of effects, they inherit the other weaknesses of PDPs. Moreover, they scale poorly with the number of data cases, as they tend to overplot significantly, obscuring potentially interesting curves.

%% file: sections/relatedWork.tex
\section{Related Work}

In this survey, we review only a relevant subset of the large literature on interpretable machine learning. Specifically, we focus on methods for tabular datasets, and ignore the areas of interpretability for image and text data, in which the notion of a feature is very different. We also narrow our focus to model explanations that are primarily visual.

\subsection{Global vs. Local Explanations}

As previously discussed, one of the major axes in this space divides techniques into \textit{global explanations} and \textit{local explanations}. Global explanations present a summary of the model without regard to specific instances or subsets of interest to the user. Inherently interpretable models are one type of global explanation. These models include linear regressions, decision trees, and rule lists. Users can generally comprehend these moodels by simply reviewing their internals (e.g. the coefficients and bias for a linear regression). While these have been frequently lambasted for subpar performance, a new class of inherently interpretable models \cite{lou2013accurate, ustun2016supersparse, chen2018interpretable, angelino2017learning} have emerged in recent years, promising predictive power comparable to black box models for some problems. Feature importance scores \cite{breiman2001random, fisher2018model, casalicchio2018visualizing} and feature interaction scores \cite{hooker2004discovering, greenwell2018simple, friedman2008predictive} are global explanations as well. The latter are discussed in more detail below.

A Partial Dependence Plot \cite{friedman2001greedy} is a common form of global explanation. The entire model can be summarized with a series of single line charts (one for each feature in the model). The effect communicated is an aggregate behavior that may not represent the prediction process for any specific instance. ICE plots could perhaps be nominally categorized as a local method, since they bind one encoding (a curve) per data point. However, in practice, overplotting obscures many of the points, and no prior work has provided utilities for a user to inspect a single point's ICE curve. Therefore, they are more accurately viewed as a global explanation that provides some additional information over PDPs.

Many visual analytics systems for model analysis and debugging (see the excellent survey in \cite{hohman2018visual}) employ model summaries as one of the available views. While these systems tend to focus on the internal elements of neural networks or other specific model types, these overviews are another type of global explanation.

Local explanations focus on the prediction for a single data case. The major use cases for these approaches include consumer-oriented applications (why was my loan application denied?) or model debugging. Furthermore, these explanations mirror the techniques humans use to explain causality to each other \cite{miller2017explanation}.

Local explanations are sometimes communicated using a counterfactual; for example, ``the prediction for this instance would move from negative to positive if feature X changed by Y\%.'' One method for developing these explanations is the Growing Spheres algorithm \cite{laugel2017inverse}, which identifies the nearest dissimilar prediction in the data space and generates an explanation from the differences in the two points. Prospector \cite{krause2016interacting} uses partial dependence curves to allow users to interactively generate synthetic data points that serve as counterfactuals.

Another class of local explanations uses prototypical examples of correct and incorrect classifications to explain a model \cite{kim2016examples}. This and other exemplar-based approaches do not provide an explanation per-se, but rather operate on the premise that an explanation will be relatively clear to a subject matter expert once the examples are surfaced (e.g. they will note that all of the incorrect classifications had a particular unusual value for a certain feature). 

The most well-known local approach is the LIME algorithm \cite{ribeiro2016should} which fits a model on a set of points randomly drawn from a Gaussian distribution centered on the dataset's mean. Intuitively, this builds a model that captures the impact of slight movements around the data space centered on an instance of interest. The resulting model (typically linear) is sparse and interpretable.

\subsection{Regional Explanations}
There are clear downsides to both global and local approaches. Global approaches by definition sacrifice complexity and fidelity to the original model for simplicity. At the same time, local models tend to only be appropriate for specific use cases - a data scientist could not realistically debug a model by generating LIME explanations for 10 random instances out of a dataset of 100,000 records. In other words, existing local approaches provide no indication as to how they generalize beyond the instance in question.

VINE falls under the category of \textit{regional explanations}, a novel category description under which we believe several pieces of prior work can be fruitfully categorized. Regional explanations split the difference between global and local approaches by describing behaviors that affect significant regions of the data space. This affords more generality than local approaches, but yields more specificity than global approaches. Regional explanations can be thought of as exceptions to global behavior - the global behavior (e.g. a PDP line) applies \textit{unless} a data case falls in a specific cluster. We define regional explanations as explanations that meet at least one of the following criteria:

\begin{itemize}
    \item \textbf{\textit{C1}} An algorithm identifies a region of the data space in which many of the points share a common behavior in the model. A succinct description of this cluster is provided.
    \item \textbf{\textit{C2}} The common behavior for this data space is described.
\end{itemize}

Below, we review related work that qualifies under this definition.

\subsubsection{Subset-selection-based approaches}

Many visual analytics systems provide utilities for users to select an arbitrary subset(s) of interest either by predicate or direct manipulation. Users can then compare outcomes such as accuracy, or model internals such as nodes in a neural network. The GridViz application was developed by Google to help them understand a model for advertising click predictions by visually comparing slices of the data \cite{mcmahan2013ad}. MLCubeExplorer displays a wide variety of distribution, prediction, and correlation data about subsets, with the intent of comparing the relative values of two models \cite{kahng2016visual}. ActiVis \cite{kahng2018cti} allows a user to select instances of interest from a visualization of model results, and compare them in a ``neuron activation matrix'' view that can surface common activation channels.

While these approaches meet criteria \textbf{\textit{C2}} above, they do not meet \textbf{\textit{C1}}, as the subsets are not algorithmically generated. While interactive cohort construction is undoubtedly a useful tool, we argue that these approaches do not extract subsets \textit{which the model itself treats differently, and which may or may not correspond to human intuition}.

\subsubsection{Rule-based approaches}

A wide variety of classifiers use a system of rules to make or explain predictions. Often, these rules take the form of a predicate (if feature X $<=$ a, then predict positive). In this section we focus on rule-based methods that are specifically engineered for providing explanations - we do not consider a 10-layer decision tree interpretable by the average human.

One class of rule-based approaches consists of inherently interpretable models that use a series of rules to make a prediction \cite{angelino2017learning, letham2015interpretable, friedman2008predictive}. RuleMatrix \cite{ming2019rulematrix} uses a similar approach to generate rules that describe an underlying model, then presents the rules in an interactive visualization. Anchors \cite{ribeiro2018anchors} are model-agnostic explanations that use high-accuracy rules to define model behaviors. If a data case meets the criteria defined by a predicate, then it is highly likely that the anchor's predicted value holds, regardless of the values of the other features of the data case. However the rules extracted by this algorithm do not cover much of the data space.

While these rule-based approaches clearly define regions (\textbf{\textit{C1}}), the behavior that they define for the region is very coarse, consisting of a single value (the prediction). 

\subsubsection{Clustering Approaches}

Explanation Explorer \cite{krause2017workflow}, Rivelo \cite{tamagnini2017interpreting} and related tools \cite{krause2018user} generate local models for each datapoint, consisting of a minimal list of features that would strongly affect the prediction if changed. Datapoints are aggregated into clusters based on having identical or similar local models. Users can then view the details of instances in the cluster, as well as their evaluation metrics (e.g. the number of points predicted for each class, accuracy, etc). \cite{krause2016using} presents Class Signatures, which expand on these methods by clustering instances by feature importance lists AND prediction, thus creating more nuanced groups. 

These tools deal exclusively with binary features and a binary target - the data type can be either tabular or text. This approach defines regions (\textbf{\textit{C1}}) of the dataset, but due to the nature of binary features, there is less need to describe behavior (\textbf{\textit{C2}}) for the cluster. The authors note that their approach is more fine-grained than feature importance scores. While this is true, tabular datasets with numerical and ordinal features require more complex expressions of behavior, for which partial dependence curves are well-suited.

Shapley Additive Explanations (SHAPs) \cite{lundberg2018consistent} leverage a well-established game theory method to generate feature importances \cite{vstrumbelj2014explaining}, and extend this technique to include variables representing feature interactions. These variables are then combined into an additive explanation for each point in the dataset. While this explanation is not sparse on its own, it allows instances to be clustered based on the ordering of feature importance values. The authors annotate their visualizations with hand-curated labels for clusters that are found to correspond to shared real-world explanations (e.g. these data points were predicted to have low income because they are young and single). However, it should be noted that while SHAPs automate the identification of regions (clusters), they do not algorithmically generate sparse explanations for these clusters. Moreover, VINE provides granular visualizations of the interaction behavior due to our use of ICE curves, whereas SHAPs are not primarily a visualization tool.

\subsection{Feature Interactions}

Regional explanations can also be understood as a form of statistical interaction effect between two features, when the effects of two features upon a prediction are non-additive. Prior work has primarily focused on quantifying the strength of these non-additive relationships via interaction scores. While the literature uses a variety of terms for these effects, such as statistical interactions, feature interaction effects, and non-additive interactions, we use the term \textit{feature interactions} throughout.

\subsubsection{Model Types that Inherently Explain Interactions}

Several types of GLMs (Generalized Linear Models)  include features that model interaction effects. RuleFit \cite{friedman2008predictive} is a modified linear regression that includes interaction terms which are derived from the splits generated by tree ensembles. This method for generating interaction terms, and subsequent pruning with a regularization algorithm, ensures that RuleFit models are relatively sparse.

Another type of GLM is GAMs (Generalized Additive Models). A GAM is essentially a linear model in which each feature can be modified by a \textit{link function} which enables the model to capture non-linear (say, logarithmic or quadratic) relationships between the feature and the target. A modified version, GA$^2$M's, adds interaction terms consisting of two features, which are again modified by a link function \cite{lou2013accurate}. The GAMUT visual analytics system \cite{hohman2019gamut} uses GAM curves (as well as instance-based explanations) as a model explanation tool, in much the same way as partial dependence curves.

Both GA$^2$Ms and RuleFit present clear explanations for individual features, but suffer from the difficulty of interpreting interaction terms. GAMs are incapable of modeling feature interactions.

\subsubsection{Score-based}

One method for measuring interaction strength is the H-statistic \cite{friedman2008predictive}, which compares the 2-D partial dependence function for two features against the sum of the individual partial dependence functions for each feature. The loss is used to generate the interaction score, as it captures the degree to which additive explanations fail to recapture the target. Partial dependence functions have also been leveraged to calculate feature interactions \cite{greenwell2018simple}. This method observes the partial dependence function for feature A at various intervals of feature B, and calculates the variance in the PD function across all points. Intuitively, this method treats features A and B as independent if feature A's importance to the model remains constant regardless of feature B's value. While these methods generate numerical scores, the authors of their respective papers choose to communicate the scores with simple graphics, such as bar charts. Arguably, this is a natural mode of expression for this data. 

These methods only identify the presence and strength (in terms of average impact on a prediction) of feature interactions. They do not indicate regions of a feature's range in which interactions might be particularly strong or weak or the shape of the function that expresses the interaction.

\subsubsection{Visual}

In a Variable Interaction Network (VIN) \cite{hooker2004discovering}, features are displayed in a stylized network graph in which connections indicate the presence of an interaction. This method is notable for its ability to efficiently identify interactions including 3 or more terms. The interactions are identified by an algorithm that uses a permutation method similar to feature importance scores \cite{breiman2001random} to identify features whose effect changes in the presence or absence of a potential interactor feature. The algorithm then cleverly prunes the search space by using the property that an interaction effect can only exist if all the lower-order effects that involve its feature also exist. Similar to the H-statistic, Variable Interaction Networks do not communicate granular detail about the nature of interactions, only their presence.

ICE and PDP plots can be extended to communicate feature interactions, in ways which leverage their visual properties but do not generate interaction scores directly. \cite{friedman2001greedy} suggests a heatmap partial dependence plot, in which color is encoded as the average predicted value for all points in the 2-D space defined by two features. This method visualizes feature interactions as color artifacts, such as sharp gradients or large areas with no variation (see for example \cite{molnar2019interpretable}). Similarly, ICE plots can encode a second variable as the color of a line \cite{goldstein2015peeking}. The most simple effect would be a correlation between hue and Y-value which would indicate that two features have a positive super-additive interaction effect. 

Partial Importance (PI) plots and Individual Conditional Importance (ICI) plots \cite{casalicchio2018visualizing} operate much as PDP and ICE plots but visualize feature importance instead of prediction value. This is a regional approach in the sense that it visualizes the regions of a feature's range in which it impacts predictions. The authors note that high variance between individual curves in an ICI plot suggests the presence of feature interactions.

ALE plots \cite{apley2016visualizing} are a solution to the aforementioned tendency of PDPs to generate inaccurate curves where features are highly correlated. ALE plots instead calculate partial dependence from small piecewise segments consisting of points with values in a narrow range, removing the need for synthetic data. These plots address the issue of feature interactions by allowing the user to view the feature's main effect, and any interaction effects in separate plots.

\subsection{Summary}

The major downside to all of these approaches is that they require significant user time and skill, and there is no predefined threshold for a ``significant'' feature interaction. A data scientist would likely need to generate a scatterplot matrix of all possible feature combinations, or try one by one, perhaps with interaction scores or a VIN as a pruning mechanism. While these methods have enormous value in the process of exploratory data analysis, they are less suited for effectively communicating model properties. 

Our approach to interaction effects is to present them when they serve as a relevant explanation for model behavior. We split the difference between relatively coarse interaction scores, and complex charts. VINE is therefore not a pure feature interaction score and does not directly compete with measures such as the H-statistic. Rather VINE curates a selection of feature interactions that aids the interpretation of model behavior by providing exceptions to global behavior. A user of VINE interprets a model using the global explanation (a PDP curve) except where a data case meets certain criteria (a region in which feature interaction effects occur). We argue that this is a parsimonious but powerful explanatory technique capable of communicating both feature interactions and non-linear relationships while not overwhelming the user with many instance-level details.

%% file: sections/approach.tex
\begin{figure}
\centering
\includegraphics[scale=0.5]{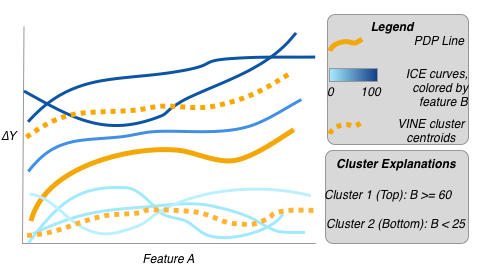}
\caption{VINE cluster curves overlaid on a plot with PDP and ICE curves to show how clusters capture regional behavior. ICE curves are colored according to an interacting feature.}
\label{fig:clustering-process}
\end{figure}

\section{Approach}

Our approach is to create a visualization for model explanation that leverages modified ICE plots and to present these plots in a visual analytic tool called VINE. We generate VINE curves via the following steps:

\begin{algorithmic}

\For{feature F in Features}
\State Cluster data using ICE curve slopes as a feature representation
\State Generate a predicate for each cluster using a 1-layer decision tree
\State Merge clusters with similar explanations
\EndFor
\end{algorithmic}

An example of this algorithm is presented in Figure \ref{fig:clustering-process}. We believe that this process produces accurate regional explanations for model behavior in the form of partial dependence curves which apply to a subset of the dataset.

\subsection{Calculating Clusters}
To address the issue of overplotting on ICE curves, VINE tries to cluster similar curves and visualize a centroid curve instead. Note that this is a form of unsupervised clustering on the dataset, but that instead of using an instance's feature vector as its representation, we instead use the X,Y tuples that constitute its ICE curve. We assessed a variety of clustering algorithms and distance metrics with the goal of generating accurate clusters quickly. Accuracy was initially assessed by visually comparing the centroids against the constituent ICE curves to validate that clusters were cleanly separated. In particular, we assessed the following clustering metrics, using implementations from scikit-learn \cite{scikit-learn}: DBSCAN, K-Means, Affinity Propagation, Agglomerative Clustering, and Birch. We found that Agglomerative Clustering \cite{ward1963hierarchical} and Birch \cite{zhang1996birch} both performed acceptably, with Birch running approximately 2.5x faster, but producing less cleanly defined clusters. Agglomerative Clustering is used for all examples in the paper, although Birch can be selected as an option when running the script.

A more difficult question was the choice of distance metric for calculating the pairwise distance between ICE curves. Euclidean distance produced groups of curves which a human would clearly recognize as inappropriate. While Dynamic Time Warp produced clusters that appeared highly appropriate to the eye, we were unable to identify a fast implementation. This was necessary because all pairwise distances must be calculated, meaning that our algorithm scales in $O(K*N^2)$ time, where \textit{K} is the number of features and \textit{N} is the number of items in the dataset. We also tried the Slope Similarity algorithm, which compares the Euclidean Distance between the slopes of ICE curves instead of their raw points.  The Slope Similarity measure produced appealing results as well, and and ran in the same time as Euclidean Distance, making this an ideal choice for our purposes.

\subsection{Generating Cluster Explanations}
After clustering the ICE curves we try to provide a human-interpretable explanation for each cluster of curves-- that is, what do these clustered curves have in common that differentiate them from the rest of the ICE curves? To answer this question, we used a 1-deep decision tree to predict membership in that cluster against all other points (one-vs-all).

This simple model identifies the feature and split value that most reduces the entropy between the curves in the cluster and those outside of the cluster. Intuitively, this split represents a good explanation for what characteristics make the cluster unique.

\subsection{Merging Clusters}

One difficulty with our method was in choosing the appropriate number of clusters for each dataset. We simulated exploratory data analysis (EDA) with early versions of the tool and found that some features would produce 5 or more distinct clusters of behavior (itself an interesting result), but that for other features, many of the cluster explanations would be duplicative, or nearly so (e.g. two clusters with the explanation $Weight>3$). All else being equal, it is preferable to have fewer clusters so as to reduce the visual complexity of the chart and to allow users to focus on a few highly salient behaviors. To prune the list of explanations, we chose to implement a cluster merging operation, given the lack of any a priori indicator for the ideal number of clusters. In practice, we noticed that the accuracy of a merged cluster is usually higher than the mean accuracy of two clusters with similar explanations. 

Merging was accomplished via the following process:

\begin{algorithmic}
    \For{cluster $c_i$ in $C_1$ to $C_N$}:
    \For{cluster $c_{dupe}$ in $C_{i+1}$ to $C_N$}:
    \If{$c_i.feature = c_{dupe}.feature \land c_i.direction = c_{dupe}.direction$}
    \If{$\frac{(c_i.{value}-c_{dupe}.value)}{f_max-f_min} <= 0.05$}
    \State Merge Clusters
    \EndIf
    \EndIf
    \EndFor
    \EndFor
\end{algorithmic}

Here, each cluster's explanation has a ``feature'' property (the feature used to define the split), a ``direction'' property ($<=$ or $>$) and a value property (e.g. 3) which together define a predicate ($Weight <= 3$). \textit{f} is the feature for which the plot is being generated.

%% file: sections/implementation.tex
\section{Implementation}

Our algorithm was built in Python 2.7, using standard machine learning libraries, including Numpy, Pandas, Scipy, and Scikit-Learn \cite{scikit-learn}. In addition, the original code for calculating PDP and ICE curves was forked from the PyCEBox library \cite{austin_2018}, though it has been heavily modified in our implementation. We also employed the sklearn-gbmi package \cite{haygood_2017} to calculate H-statistics. The charts in the paper were generated with Altair \cite{Altair2018} and Matplotlib \cite{HunterMatplotlib2007}. The VINE visual analytics system was built in HTML using D3.js \cite{2011-d3}. It consumes a JSON file that is output by the Python script.

VINE initially presents the user with a feature space visualization designed to communicate the relevance of each feature to the model (see Figure \ref{fig:teaser}). VINE charts as presented as small multiples, one per feature. The X-axis indicates the strength of feature interactions. The Y-axis indicates the overall feature importance. This allows the user to quickly familiarize themselves with the dataset and its salient features. Both the charts themselves and their position in the feature space draw the eye to interesting patterns. For example, in Figure \ref{fig:teaser}, \textit{Hour of Day} is clearly the most important (topmost) feature, which can be verified by checking its Y-axis scale. \textit{Work Day}, on the bottom right, is not a particularly important feature most of the time, but it does have one interaction (the red bar) which produces an outsize effect. Because this effect is so different than the PDP term, \textit{Work Day} has a strong feature interaction score and occupies a sparse corner of the feature space. \textit{Wind Speed}, on the other hand, has no VINE curves at all, and so appears on the left side of the graph.

The feature interaction strength (X-axis) is calculated as the sum of Dynamic Time Warp distances between each VINE curve and the PDP curve, normalized by the maximum value of the PDP curve.  Feature importance (the Y-axis), is determined by the standard deviation of the PDP curve. The position should be taken as a rough approximation, as a force layout is used to prevent overplotting of the small multiples.

Users can select a feature to enlarge the chart, which makes the explanations visible. VINE charts are displayed in the same manner as PDP and ICE plots. The VINE chart for feature A will have feature A's range as the X-axis. The Y-axis depicts the change compared to the mean prediction. We chose to mean-center each plot to enable an additive interpretation, i.e. for a given data point, a user would sum the values from each plot to arrive at a prediction, rather than the traditional PDP which requires the values to be averaged. The partial dependence curve is presented as a black line. Each colored line represents a VINE cluster and is calculated as the centroid of all its constituent ICE curves (in other words, a partial dependence curve for the subset). The width of each VINE curve encodes the size of its cluster, but is log-scaled for readability purposes. Clicking on a VINE curve reveals all its constituent ICE curves. This allows the user to visually inspect the quality of each VINE curve. 

Binary features are presented as bar charts instead of lines, to aid in their interpretation and visually distinguish them from numeric features. However, the underlying VINE algorithm is applied identically to each feature. The bar charts use the same color scheme as the line charts, with black corresponding to the PDP. A bar should be interpreted as the change in prediction incurred by increasing a feature from 0 to 1.

Lastly, the histograms on the right side provide a visual depiction of the explanation for each cluster. The histograms can be mapped to a VINE curve based on color. One or more columns will be displayed depending on the number of features that appear in explanations. In Figure \ref{fig:use-case-2}, the \textit{Hour of Day} feature serves as the best explanation for two of three VINE curves. The darker green region of the histogram conveys both the range defined by the explanation and the density of points in that region. The text of the definition, the size of the cluster and its accuracy are displayed in the top right-hand corner of the chart.

%% file: sections/evaluation.tex
\section{Evaluation}

We evaluated the VINE algorithm on three benchmark tests, including an application of the Information Ceiling framework. Each test was performed on three datasets using a regression target and a single model fit for this task.

\subsection{Datasets}

VINE was evaluated on three tabular datasets with numerical, ordinal, and categorical features. Pre-processing consisted of one-hot encoding any categorical features. Ordinal features, such as \textit{Month} for the Bike dataset, were left as is. These datasets did not have missing or erroneous values and so no imputation was performed. Due to the choice of a tree-based model, normalization/standardization was not necessary. The version of the Bike dataset stored in the UCI repository has several standardized features - these were transformed back to their original domain for readability purposes. Several features were removed from the Bike dataset \cite{dua2017UCI} in order to produce a more intelligible model. Weekday and holiday were removed because they were raw versions of the engineered \textit{Workingday} feature. \textit{Dteday} and \textit{Month} were removed for similar reasons, because they were better represented by the \textit{Season} feature. The \textit{Casual} and \textit{Registered} variables were removed because they are alternate regression targets, and highly correlated with the \textit{Cnt} target. Feature names for the Bike dataset have been changed to make them more human-readable for figures and use cases in this paper.

For all datasets, a Gradient Boosting Regressor was used. Each regressor used 300 trees and a minimum leaf size of 100 to prevent overfitting. The accuracy of each classifier is generally high and is reported in Table \href{tab:datasets}. Hyperparameters were manually selected to produce decently accurate classifiers that were faithful to the underlying dataset, but beyond these basic measures, no attempts were made to identify an optimal model. For our purposes, the model is of more interest than the dataset or the relationship between the two. For this same reason, the entire dataset was used to fit the model, as there is no use for a test set in our evaluation. That said, VINE should perform similarly on unseen data as long as it was drawn from the same distribution as the data used to fit the model.

Note that for all three datasets, we chose a regression problem as our task. However, we believe that binary or multi-class classification problems can easily be tackled with VINE as well, as PDP and ICE curves are also suitable for these tasks. The major difference for these tasks is that the interpretation of the Y-axis alters from ``change in prediction'' to ``change in probability of a given class''. 

\begin{table}[tb]
  \caption{Datasets used in our evaluation. Bike dataset is available at \cite{dua2017UCI}. Other datasets were loaded from scikit-learn \cite{scikit-learn}}
  \label{tab:datasets}
  \scriptsize%
	\centering%
  \begin{tabu}{%
	l|c|c|c|l
	}
  \toprule
   Name & Features & Instances & Model $r^2$ & Target \\
  \midrule
  Diabetes & 10 & 442 & 0.912 & Disease severity. \\
  Boston Housing & 13 & 506 & 0.987 & Median house price. \\
  Bike Sharing & 11 & 17,379 & 0.907 & Hourly bike rentals. \\
  \midrule
  \bottomrule
  \end{tabu}%
\end{table}

\subsection{Comparison to Random Clustering Baseline}

We first attempted to evaluate the efficacy of our algorithm for generating clusters and their corresponding explanations. We sought to ensure that our cluster explanations were accurate and that they outperformed a nominal baseline approach. The purpose of this check was to demonstrate that our decision tree would not simply overfit random clusters to a nonsensical explanation, and that a real signal must be present in the cluster in order for a high-accuracy explanation to be generated.

To evaluate the explanations, we compared the data points contained in each cluster (set A) with the data points returned by filtering the dataset on the cluster explanation (set B). By treating set A as a training set and set B as the model output, we were able to apply traditional accuracy, precision, and recall metrics. For this evaluation, we set the hyperparameter for number of clusters to 5.

To generate a baseline comparison, we used the following method to generate random clusters:

\begin{algorithmic}
    \For{feature $f_i$ in $F_1$ to $F_N$}:
        \State Partition the dataset into 5 clusters of random size
        \State Fit a decision tree to each cluster, as in the VINE algorithm.
        \State Calculate the accuracy, precision, and recall as discussed above
    \EndFor
\end{algorithmic}

\subsection{Correspondence to H-statistic Results}

We believe that the explanations our method returns should be consistent with existing methods that quantify feature importance and feature interactions. Assuming that feature A interacts strongly with features B,C,D according to a measure such as the H-Statistic \cite{friedman2008predictive} or Greenwell's partial dependence interaction \cite{greenwell2018simple}, then we expect to see that cluster explanations for feature A will include feature B, C, and/or D, allowing for the possibility that other features may be included as well, due to the fact that an interaction may only be strong in a narrow range. 

To this end, we evaluate our cluster explanations using Friedman's H-statistic, which generates a score between 0 and 1 for each pair of features. 0 indicates no interaction between the two features, and 1 indicates that the features have no main effects, but rather that their entire impact on the prediction is generated from their interaction. For a given Feature A, we generate a list of features that appear in its cluster explanations (list A). We compare list A against the list of feature interactions, ordered by the H-statistic (list B). 

Given that one of the issues with the H-statistic is the lack of a well-established threshold for determining significance, we chose to ignore the values themselves and instead calculate the number of elements from list A that appear in the top 3 features of list B. We then sum this count across all features in the model, and normalize it by the total number of clusters generated by VINE. The result can be interpreted as the percentage of explanations that utilize a strongly interacting feature. We also present the baseline probability that features would have appeared among the top 3 interactors if they were chosen at random (this probability is constant for each dataset, equal to $\frac{3}{number_of_features}$).

\begin{figure}
\centering
\includegraphics[scale=0.2]{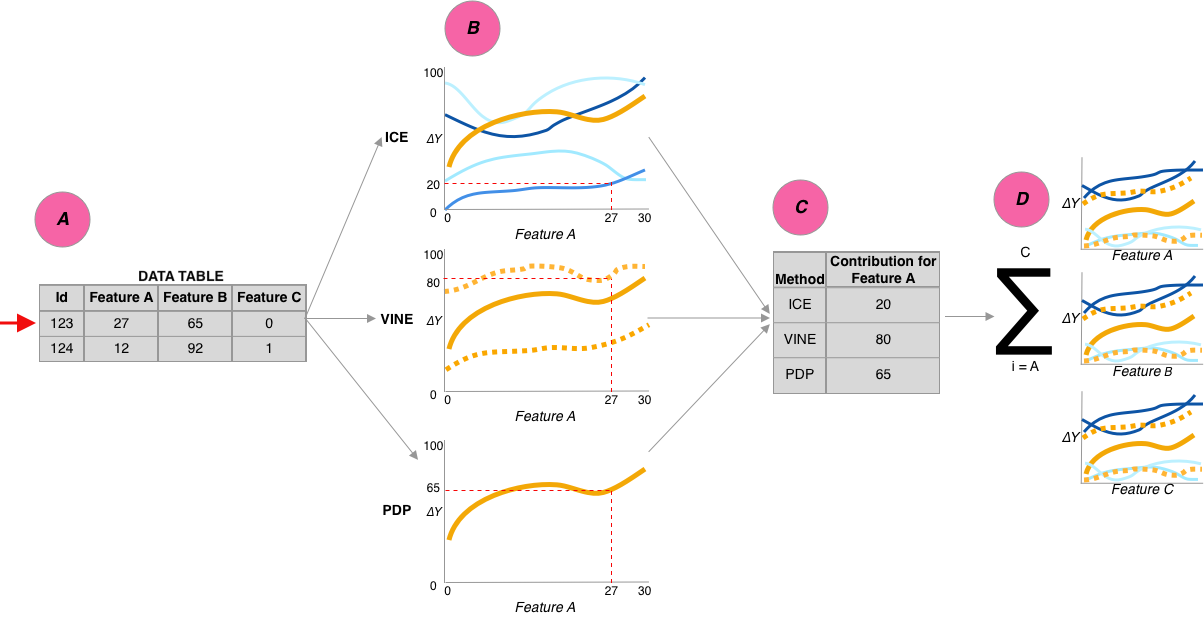}
\caption{The process for calculating the Information Ceiling metrics for PDP, VINE, and ICE plots. (A) Generate predictions for one point at a time. (B) Identify the relevant curve. For VINE, select the curve whose predicate includes the point. For ICE, find the ICE curve corresponding to that data point. (C) Get the $\Delta Y$ value from the appropriate curve at the point's feature value. (D) Sum the results of steps A-C for each feature. This is added to the mean prediction and compared to the model prediction to find the loss.}
\label{fig:information-ceiling-process}
\end{figure}

\subsection{Information Ceiling}

We introduce a novel framework, the Information Ceiling, for evaluating the fidelity of any visual model explanation to its underlying model. For our tabular regression problems presented here, the metric simply consists of the $r^2$ (often known as the Coefficient of Determination) between the model's predictions and our algorithm's predictions \textit{as it tries to simulate the human sensemaking process afforded by the model visualization in question}. The tricky part here is to describe and systematize a process by which a consumer of a visualization would use it to make a prediction. Nonetheless, as this is one of the most common human tasks used to evaluate visualizations \cite{doshi2017towards}, we argue that it behooves the designer of model visualizations to build them according to standard human-computer interaction principles, with specific tasks in mind.

Luckily, for VINE curves and other plots in the PDP family, a fairly simple method presents itself for making predictions based on the explanation. For the PDP, the chart for Feature A allows the user to identify the value contributed to the prediction at any point on the X-axis (e.g. the range of Feature A). To find this component of a prediction for instance X, a user simply has to find instance X's value for Feature A, find that point on the X-axis, and follow it up to the PDP line. This will yield Feature A's contribution to the prediction. The user can simply sum the results of this process for each feature in the dataset, add the sum to the mean value of the target variable, and yield a prediction based on the PDP curve. This process is summarized in Figure \ref{fig:information-ceiling-process}.

While it is unlikely that a user would perform this exact task in practice, a heuristic version is more likely. A user would notice that an instance of interest has high values for Features A,C, and D, and remember that the PDP curved sharply upwards for Features A and C. The user would add some estimated amount to an average value for the target, and produce a prediction in this manner. This method is recommended in \cite{hooker2004discovering} as a workflow for data scientists when using partial dependence plots to analyze a model.

This method can easily be extended to ICE and VINE plots, as summarized in Figure \ref{fig:information-ceiling-process}. For ICE curves, the user simply selects the particular curve for the instance of interest instead of a PDP line. For VINE, they select (much more easily) the VINE curve whose predicate matches their instance. For VINE, two edge cases must be considered: (1) when a point matches 2 or more predicates, we take the mean of each of their predictions, and (2) when a point doesn't match any predicate, we use the PDP line for prediction instead.

This method is easy for an algorithm to simulate when presented with the data that underlies each of the curves. It should be noted that we do not expect any user to derive predictions as accurately as our algorithm can. Instead, we treat our metric as the upper limit on prediction fidelity (or a lower bound on error) that could possibly be achieved by interpreting the visualization in this way. For this reason, we refer to this evaluation framework as the Information Ceiling.

%% file: sections/results.tex
\section{Results}

We report performance on three algorithmic benchmark tests across three datasets. Each of the benchmarks was devised specifically for this paper. A Jupyter notebook with the full code required to reproduce all results, charts, and tables in this publication is available at \url{https://www.github.com/MattJBritton/VINE}. Instructions, code, datasets, and other files necessary to run VINE as a standalone tool are also available at this URL.

\subsection{Comparison to Random Clustering Baseline}

\begin{figure}
\centering
\includegraphics[scale=0.35]{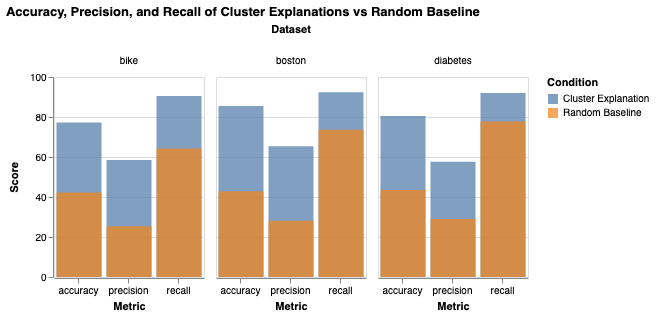}
\caption{VINE cluster explanations are far more accurate when generated on real as opposed to random clusters.}
\label{fig:cluster-accuracy}
\end{figure}

Figure \ref{fig:cluster-accuracy} indicates that VINE cluster explanations more accurately describe real subsets than randomly chosen subsets. We take this as evidence that VINE explanations detect real descriptions of subsets, and do not simply fit noise.

\subsection{Correspondence to H-statistic results}

\begin{table}[tb]
  \caption{Explanations used in VINE tend to have high H-statistic values}
  \label{tab:h-statistic}
  \scriptsize%
	\centering%
  \begin{tabu}{l|c|c
	}
  \toprule
   Dataset & \% in Top 3 & Baseline \% with Random Assignment \\
  \midrule
  Diabetes & 60.5\% & 30\% \\
  Boston Housing & 64.2\% & 23.1\% \\
  Bike Sharing & 65.7\% &  27.3\%\\
  \midrule
  \bottomrule
  \end{tabu}%
\end{table}

Table \ref{tab:h-statistic} presents the results of the H-statistic experiment. The feature used in VINE explanations occurs in the top 3 interactors (sorted by H-statistic) about twice as often as we would expect it to if features were selected randomly. This suggests that the VINE algorithm successfully measures feature interactions. Note that the H-statistic calculation (and the random baseline) are non-deterministic, so results will vary across iterations. Results for one pass are reported.

\subsection{Information ceiling}

\begin{figure}
\centering
\includegraphics[scale=0.3]{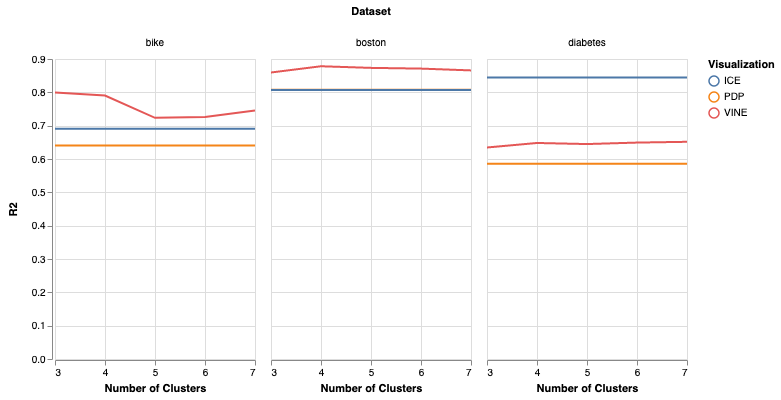}
\caption{Information Ceiling of PDP, ICE, and VINE plots by dataset. VINE consistently outperforms PDP.}
\label{fig:information-ceiling-data}
\end{figure}

Our Information Ceiling method shows that VINE curves have higher fidelity to the model than PDPs (see Figure \ref{fig:information-ceiling-data}). In addition, our method outperformed Individual Conditional Expectation plots in two of the three datasets. We conclude that our method can be considered a more accurate representation of a model's behavior than PDPs. In addition, it appears that the method for calculating individual conditional expectation has fundamental limitations, which may be caused by the aforementioned issues with extrapolation. Even when a prediction is generated for a data point based on its own ICE curve, the prediction is scarcely better than the PDP line (for two of the three datasets). We hypothesize that when VINE aggregates ICE curves, it averages out instabilities, which is ample tradeoff for the loss in specificity. 

%% file: sections/discussion.tex
\section{Discussion}

\begin{figure}
\centering
\includegraphics[scale=0.26]{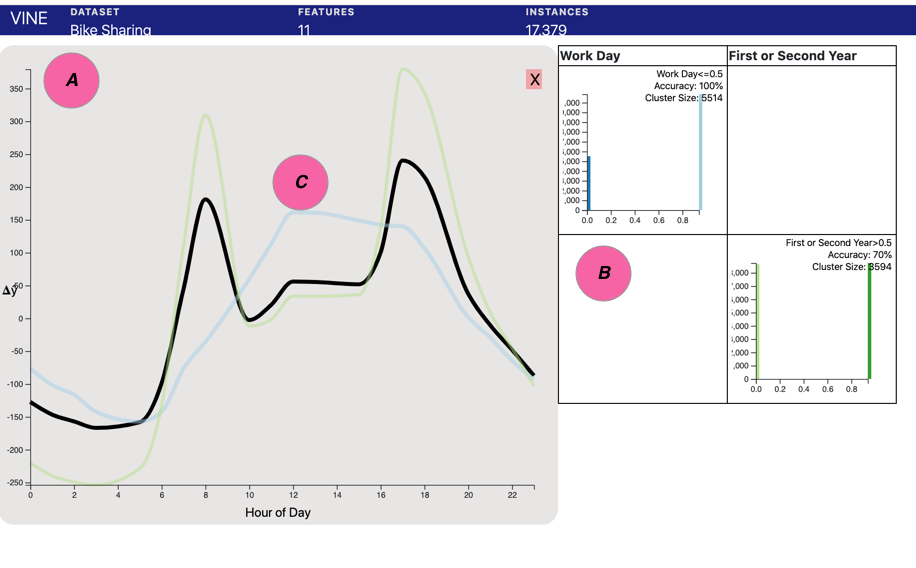}
\caption{VINE plot for the \textit{Hour of Day} feature in the bike dataset. (A) The main plot shows the PDP as a black line, and VINE curves as various colors. (B) The sidebar uses matching-color histograms to visualize the explanation for that subset. (C) The blue curve visualizes an important insight - the typical rush hour peak pattern does not exist on weekends.}
\label{fig:use-case-1}
\end{figure}

\subsection{Contributions}
Our contribution consists of (1) an algorithm that clusters ICE curves based on shape similarity and generates a human readable label for that subset, (2) a visual analytics tool that facilitates model interpretation and sensemaking using VINE explanations, and (3) a framework for evaluating visual explanations of machine learning models based on the loss that an automated method incurs when using them as a basis for prediction.

\subsection{Strengths}
(1) Our algorithm is completely model-agnostic. The only requirement is that the model's prediction function be passed into the \textbf{export} method and that this prediction function uses the same API as scikit-learn \cite{scikit-learn}.(2) VINE curves extract salient feature interactions and give detailed information about how they affect predictions. Identifying these feature interactions is as simple as reading the chart, and does not require a detailed statistical analysis.(3) The Information Ceiling framework allows us to compare the validity of multiple visualizations in the partial dependence family for the first time.

\subsection{Limitations}
(1) Our approach is currently limited to tabular data and does not work for text, image, or video data. (2) Our approach works best when most features in a dataset are numerical. Ordinal, categorical, and Boolean features are supported, but existing methods \cite{krause2016using, krause2018user, krause2017workflow, tamagnini2017interpreting} are better adapted to this task. In particular, one-hot encoding a categorical variable or creating a vectorized text representation can create a confusing array of features. (3) Large datasets ($>$50,000 rows) will take at least several minutes to compute and may use a large amount of memory.

\subsection{Potential Use Cases}
Our tool extracts model behavior that differs significantly from the mean feature effect (the partial dependence curve). This has enormous potential value for debugging both the model and the training set. We used an early version of our tool to perform some model debugging on the \textit{Month} attribute of the bike dataset. Data cases with a \textit{Season} of Spring and a month from July-December had markedly lower predicted ridership than the PDP average. However, this combination of features (Spring in December) is impossible. A data scientist could take this insight and either build a validation rule for data intake, or more likely drop one of the two highly correlated features.

Another use case is the extraction of insights. Our tool can partially automate or supplement exploratory data analysis. In Figure \ref{fig:use-case-1}, an analyst viewing the \textit{Hour} feature in the bike dataset would note that VINE has found two regions of interest. The blue VINE curve is for weekends/holidays (\textit{Workingday}=0). The PDP curve shows a large bump in ridership at the morning and evening rush hours. However, for weekends, this effect is far less pronounced, with ridership increasing steadily but less sharply. The insight that weekday and weekend ridership patterns are fundamentally different is presumably valuable for a bike-sharing company. These insights are extracted without the user being aware of the importance of the \textit{Workingday} feature or making any intentional effort to analyze it. 

Figure \ref{fig:use-case-1} visualizes the effect of the \textit{Feels Temperature} (temperature + wind chill) on ridership. The PDP curve indicates that the model predicts a large spike in ridership around 75 degrees. However, the VINE curves reveal a more nuanced story. Later afternoon and evening ridership (the blue curve) spikes higher, while early afternoon and morning ridership (the red curve) stays mostly flat until the temperature becomes very hot. Moreover, the green curve indicates that on warm winter days, ridership spikes particularly high and at a lower temperature. A model explanation communicated with only the PDP curve might convince the bike-sharing company to reduce the size of the fleet during the winter. VINE flags the potential for highly profitable warm winter days. It is unlikely that this correlation would have been discovered unless someone thought to check for it explicitly.

\begin{figure}
\centering
\includegraphics[scale=0.26]{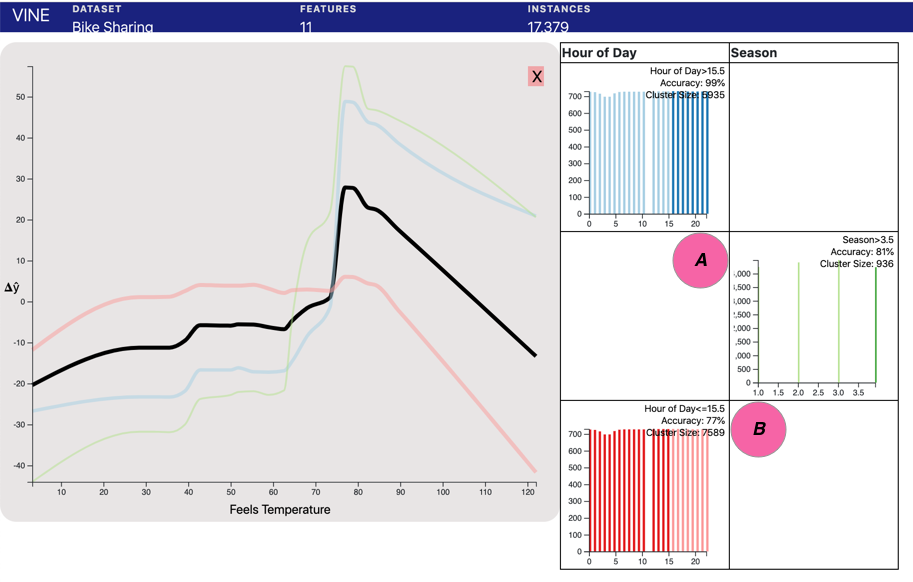}
\caption{Insights into the effect of temperature on ridership for the bike dataset. (A) The model predicts a spike in ridership around 70 degrees. This behavior is particularly strong in the winter months. (B) However, this is a clear exception to this behavior. Ridership in the early afternoon and morning does not spike.}
\label{fig:use-case-2}
\end{figure}

\subsection{Evaluation Framework}

We believe that the Information Ceiling metric can be used to validate the effectiveness of a wide array of visualizations in the interpretable ML space. While we only consider PDP, VINE, and ICE plots here, it would be trivial to compare ALE plots too. Commentators \cite{molnar2019interpretable} have noticed that whereas PDP plots suffer from extrapolation into sparse areas of the conditional distribution, ALE plots can suffer from a related tradeoff between accuracy and stability when setting the hyperparameter for number of intervals. An easy way to determine the superior method is to evaluate them using our method, which quantifies fidelity to a model.

Beyond visualizations in the partial dependence family, our Information Ceiling framework could also be used to evaluate explanations such as RuleMatrix \cite{ming2019rulematrix}, in which the algorithm would simply scan through rules in the order they are presented in the visualization until it found a matching predicate for a given instance. Similarly, Gamut \cite{hohman2019gamut} or other GLMs can be evaluated in much the same way as PDPs, essentially using a feature plot as a lookup table for each instance and then adding predictions together.

Pushing the envelope further, it should be possible to evaluate LIME \cite{ribeiro2016should}, creating a direct comparison between global and local explanations for the first time. One approach, based on our personal model interpretation workflows, is as follows: (1) generate \textit{k} cluster centroids using a method such as k-medoids, (2) build a LIME model for each centroid, and (3) make a prediction for an instance by finding the nearest centroid and using its LIME model. Clearly, there are many undefined parameters here, such as the value \textit{k} or the distance metric to use. It is likely that the human sensemaking process for this task is difficult to replicate as an algorithm. However, we argue that there is value in investigating this process, under the assumption that it will not be possible to design a good model visualization (or indeed, any visualization) without a sense of its intended use.

It should be stressed that we do not recommend evaluating explanations solely by our method. Our method is not capable of measuring the aesthetic value or ease of interpretation of an explanation, only its information content. We believe our Information Ceiling framework can instead set a ceiling on the understanding that a human can glean from an explanation. It should be noted that this is not a new concern in information visualization - prior research has investigated the fidelity of visualization-generating algorithms such as t-SNE \cite{wattenberg2016how} or even histograms \cite{lunzerhistograms} to the underlying data.

Other research \cite{narayanan2018humans} has investigated the design and perceptual factors involved in model visualization. This research is a necessary complement to our work, addressing how to ``make the most'' of the Information Ceiling that a visual model explanation affords. Performance on any prediction task performed by a human can be compared directly to the Information Ceiling, and the loss can be explained by either design issues with the visualization, or human perceptual limitations (e.g. people have been shown to exaggerate certain features of line charts and downplay or excise others \cite{mannino2018qetch}).

%% file: sections/futureWork.tex
\section{Future Work}

\begin{itemize}

\item \textbf{Investigate the effectiveness of partial dependence across datasets.} While it cannot be proven from this limited study, the far higher performance of the ICE plot on the Diabetes dataset suggests that the fidelity of partial dependence curves may be contingent on some unknown property of a dataset, such as the presence of multi-collinearity. The Information Ceiling provides an ideal tool to probe the limitations of partial dependence methods and the impact of violating their assumptions. Given the wide use of the technique, this meta-learning could be valuable.

\item \textbf{Evaluation.} We present a novel evaluation framework for model visualizations in which we seek to quantify their information content. We hope that this method can be used both to evaluate the fundamental validity of other techniques in interpretable ML and to guide future studies into human sensemaking with predictive models. We also hope that VINE can be evaluated in situ to determine its utility for data scientists.
\end{itemize}

%% file: sections/conclusion.tex
\section{Conclusion}

We present VINE, an interactive visualization that communicates regional explanations for models built to make predictions on tabular data. Our approach leverages existing work into partial dependence plots to derive groups of instances whose behavior in a given model significantly differs from the mean feature effect. These regional explanations also capture feature interaction effects in a novel manner. We argue that our approach provides a useful complement to global explanations by identifying caveats to the main behaviors, and also complements local explanations by aggregating similar instances and distilling common contributors to their prediction. Our approach has applicability to model interpretation and explanation, model debugging, algorithmic fairness and adversarial artificial intelligence. 

We demonstrate that our algorithm produces explanations that have high mean accuracy in describing relevant subsets. We also provide example use cases that demonstrate how VINE can facilitate model exploration on a real dataset. We evaluate VINE against PDPs using a novel evaluation framework (Information Ceiling) and find that VINE more faithfully replicates the predictions made by the model. We conclude by discussing ways in which the Information Ceiling approach can be used to quantify the effectiveness of visualizations in the interpretable ML space.